# Combination of Evidence Using the Principle of Minimum Information Gain

S.K.M Wong and Pawan Lingras
Department of Computer Science, University of Regina, Regina, Sask., Canada, S4S 0A2
e-mail: mwong@uregina1.bitnet, lingras@uregina1.bitnet

**Abstract**
One of the most important aspects in any treatment of uncertain information is the rule of combination for updating the degrees of uncertainty. The theory of belief functions uses the Dempster rule to combine two belief functions defined by independent bodies of evidence. However, with limited dependency information about the accumulated belief the Dempster rule may lead to unsatisfactory results. The present study suggests a method to determine the accumulated belief based on the premise that the information gain from the combination process should be minimum. This method provides a mechanism that is equivalent to the Bayes rule when all the conditional probabilities are available and to the Dempster rule when the normalization constant is equal to one. The proposed principle of minimum information gain is shown to be equivalent to the maximum entropy formalism, a special case of the principle of minimum cross-entropy. The application of this principle results in a monotonic increase in belief with accumulation of consistent evidence. The suggested approach may provide a more reasonable criterion for identifying conflicts among various bodies of evidence.

## 1. Introduction

The theory of belief functions has generated considerable interest among researchers in information science because of its ability to make probability judgments based on incomplete and vague information. The belief functions can be interpreted in terms of a mapping or a compatibility relation between two different but related sets of mutually exclusive propositions. The mapping or compatibility relation may be directly related to a person's knowledge or it may be strictly an abstract construct introduced for convenience. Such a view establishes a relationship between the theory of belief functions and the Bayesian theory of probability. The Bayesian theory contains important concepts such as the Bayes rule of conditionalization and various information measures that can be useful for making numeric judgments. It may therefore be possible to extend these concepts to the theory of belief functions using the compatibility relationships.

The most important aspect in any treatment of uncertainty is perhaps the rule of combination for updating the degrees of uncertainty by combining different bodies of evidence. The theory of belief functions uses the Dempster rule for combining two belief functions. However, with limited dependency information regarding the accumulated belief the Dempster rule sometimes provides unsatisfactory results. The present study proposes a method to compute the accumulated belief by minimizing the information gain. We will show that if the complete dependency information is known, our method is equivalent to the Bayes rule. On the other hand, when the normalization constant is equal to one and no other dependency information is available the method reduces to the Dempster rule. We will also show that the principle of minimum information gain is equivalent to the maximum entropy formalism which is a special case of the principle of minimum cross-entropy. Both the minimum cross-entropy and maximum entropy formalisms have been extensively studied in the traditional probability literature (Jaynes, 1957; Kullback and Leibler, 1951; Shore and Johnson, 1980). In fact, the arguments used in favor of these formalisms provide additional support for the proposed principle of minimum information gain. This principle allows us to incorporate available conditional probabilities in the combined belief function, which is a generalization of some previous attempts (Ruspini, 1986, Yen, 1989) to incorporate dependencies in the Dempster rule. The application of our method results in a monotonic increase in belief with accumulation of consistent evidence. This is an important condition that must be satisfied by any reasonable rule of combination (Shafer, 1976). The proposed principle also leads to a criterion for identifying conflicts between different bodies of evidence.

## 2. The Theory of Belief Functions

For completeness we summarize here some of the basic notations in the theory of belief functions.

Let $T = \{ t_1, \ldots, t_n \}$ be a finite set of all possible answers to a question. We refer $T$ as the frame of discernment or simply the frame defined by this question. The power set of $T$, written $2^T$, represents the set of all propositions discerned by $T$. A function $m : 2^T \rightarrow [0, 1]$ is called a *basic probability assignment* (bpa) which satisfies the

properties:

$$m(\emptyset) = 0 \text{ and } \sum_{F \in 2^T} m(F) = 1. \tag{2.1}$$

Proposition $F$ of $2^T$ is called a focal element of the bpa $m$ if $m(F) > 0$. The value $m(F)$ measures the belief that one commits *exactly* to a proposition $F$. The total belief committed to a proposition $A$ is given by:

$$Bel(A) = \sum_{F \subseteq A} m(F), \tag{2.2}$$

where $Bel : 2^T \rightarrow [0, 1]$ is a belief function (Shafer, 1976). Another quantity referred to as the *plausibility*, written $Pl$, is defined by:

$$Pl(A) = 1 - Bel(\neg A), \tag{2.3}$$

where $\neg A$ denotes the negation of $A$. Plausibility expresses the extent to which one fails to doubt $A$, i.e., the extent to which one finds $A$ credible or plausible.

The belief functions described above were originally derived from the concepts of upper and lower probabilities (Dempster, 1967). The upper and lower probabilities are useful for transferring the probability function from one frame $S$ to another frame $T$ using a multivalued mapping or a compatibility relation.

*Definition 2.1:*
Consider two frames of discernment $S$ and $T$. An element $s \in S$ is *compatible* with an element $t \in T$, written $s \, C \, t$, if the answer $s$ to the question which defines $S$ does not exclude the possibility that $t$ is an answer to the question which defines $T$.

Compatibility is symmetric: $s \, C \, t$ if and only if $t \, C \, s$. The compatibility relationships can be used to define the notion of implication between propositions from two different frames of discernment.

*Definition 2.2:*
A proposition $A \in 2^S$ is said to *imply* another proposition $B \in 2^T$, written $A \Rightarrow B$, if $B$ contains all the elements in $T$ that are compatible with the elements in $A$. (That is, if $A$ implies $B$, then any element $b$ of T compatible with some $a \in A$ must exist in $B$).

*Definition 2.3:*
A proposition $A \in 2^S$ is said to *exactly imply* another proposition $B \in 2^T$, written $A \Leftrightarrow B$, if $A$ implies $B$ but it does not imply any proper subset of $B$.

Consider a frame $T$ which denotes the set of possible answers to a question. We are interested in obtaining a probability function $P : 2^T \rightarrow [0, 1]$ on frame $T$ based on a given evidence. Suppose it is not possible to construct such a probability function directly, but from the given evidence we can define a frame - the *evidence frame S*. Let us assume that based on the evidence a probability function $P : 2^S \rightarrow [0, 1]$ on frame $S$ is known, and for simplicity let every $s \in S$ be compatible with at least one $t \in T$ and vice versa. The issue is how to use this knowledge about frame $S$ to compute the degrees of belief in the propositions discerned by $T$. The probability function $P$ on frame $T$ can of course be constructed from the probability function on frame $S$ using the Bayes rule of conditionalization, if the conditional probabilities required in the Bayes rule are known. In practice it may not always be possible to provide an accurate estimation of these conditional probabilities. In such situations one may use belief functions instead to measure the degrees of belief in the propositions of $2^T$.

Given the probability function $P : 2^S \rightarrow [0, 1]$ on the evidence frame $S$, one can define a function $m_S : 2^T \rightarrow [0, 1]$ for any $F \in 2^T$ as follows:

$$m_S(F) = \sum_{\{s\} \Leftrightarrow F} P(\{s\}). \tag{2.4}$$

The value $m_S(F)$ is the probability that is attributed to the union of those propositions in $S$ which *exactly imply* the proposition $F \in 2^T$. It can be easily verified that $m_S$ is a basic probability assignment (bpa) satisfying equation (2.1). We can use the function $m_S$ to compute the belief function $Bel_S$ as defined by equation (2.2):

$$Bel(A) = \sum_{F \subseteq A} m(F),$$

The belief function $Bel_S$ is useful for transferring the probability function from one frame to another distinctly different but related frame. This approach is particularly useful when the available information describing the relationship between the two frames is limited. In many instances the information about the evidence may be so vague that it is not even possible to explicitly construct the evidence frame. However, when a basic probability assignment $m_S$ to the propositions in $2^T$ is known, it is always possible to construct an *abstract* evidence frame $S$ as follows. Let $F_1, \ldots, F_n$ be the focal elements of $m_S$. Then for every focal element $F_i$ of $m_S$ there is a unique $s_i \in S$ such that $s_i \Leftrightarrow F_i$, that is, $s_i\, C\, t$ for all $t \in F_i$. This means that the number of elements in the abstract frame $S$ will be the same as the number of focal elements of $m_S$. Now the known bpa $m_S$ can be viewed as a probability function $P(\{S\})$ defined on the abstract evidence frame $S$ as:

$$P(\{s\}) = m(F). \qquad (2.5)$$

Hereafter we will represent a belief function on a frame $T$ in terms of an underlying probability function $P$ defined on a distinct frame $S$ and a compatibility relation $C$ between frames $S$ and $T$.

It should be noted here that how one defines the two frames $S$ and $T$ and the compatibility relationships between their elements is relative to one's knowledge and opinion, hence, purely epistemic. It is also understood that both the frame $S$ and the compatibility relation $C$ could be abstract constructs defined mathematically and lacking any semantics due to insufficient information. Nevertheless, these abstract constructs are useful for studying the combination rule.

**Example 2.1:** Based on an evidence, $T = \{t_1, t_2, t_3\}$ is the set of all possible answers to a given question. The belief function $Bel_S$ is defined by the the following basic probability assignment $m_S$:

$$m_S(\{t_1, t_2\}) = 0.8 \text{ and } m_S(\{t_1, t_2, t_3\}) = 0.2. \qquad (2.6)$$

Assume that the basic probability numbers for all other subsets of $T$ are zero. Since the underlying evidence frame $S$ is not specified, we will construct it as follows. Start with $S = \emptyset$. For the focal element $\{t_1, t_2\}$ we include an element $s_1$ in $S$ such that $s_1 \Leftrightarrow \{t_1, t_2\}$. Similarly, for the focal element $\{t_1, t_2, t_3\}$ we include another element $s_2$ in $S$ such that $s_2 \Leftrightarrow \{t_1, t_2, t_3\}$. Since there are only two focal elements, $S = \{s_1, s_2\}$ is the underlying evidence frame. The exact implication relationships $s_1 \Leftrightarrow \{t_1, t_2\}$ and $s_2 \Leftrightarrow \{t_1, t_2, t_3\}$ define the following compatibility relation $C$ between the elements of $S$ and $T$:

$$s_1\, C\, t_1, \quad s_1\, C\, t_2, \quad s_2\, C\, t_1, \quad s_2\, C\, t_2, \quad s_2\, C\, t_3. \qquad (2.7)$$

According to equation (2.5), given $m_S$ the probability function $P(\{s\})$ on $S$ is defined by:

$$P(\{s_1\}) = 0.8 \text{ and } P(\{s_2\}) = 0.2. \quad \square \qquad (2.8)$$

## 3. Rule of Combination

The key component of any theory of uncertainty is perhaps the rule for combining evidence.

Let $S$ and $S'$ be two evidence frames for which the underlying probability functions are known based on two distinct bodies of evidence. For simplicity, we will refer to both these functions as $P$ unless it is necessary to explicitly distinguish between the functions defined on different frames. The probability functions on $S$ and $S'$ along with their respective compatibility relations $C$ between $S$ and $T$ and $C'$ between $S'$ and $T$ define two belief functions $Bel_S : 2^T \to [0, 1]$ and $Bel_{S'} : 2^T \to [0, 1]$. The first step in combining these belief functions involves the construction of a joint compatibility relation $C \oplus C'$ which is a subset of the cartesian product of $S \times S'$ and $T$. The next step is to construct a joint probability function $P : 2^{S \times S'} \to [0, 1]$ on the set $S \times S'$. Finally, the belief function on $T$ based on the combined evidence can then be constructed using the relation $C \oplus C'$ and the probability function on the set $S \times S'$.

Whether one can construct the actual compatibility relation $C \oplus C'$ depends on the availability of relevant information. This relation can be accurately specified only by the person who originally defined the other two compatibility relations $C$ and $C'$. If the actual compatibility relation is not available, then similar to the Dempster rule one can construct the relation $C \oplus C'$ between $S \times S'$ and $T$ as follows:

$$(s, s')\, C \oplus C'\, t \text{ if } s\, C\, t \text{ and } s'\, C\, t, \text{ where } (s, s') \in S \times S' \text{ and } t \in T. \qquad (3.1)$$

We wish to emphasize that the compatibility relation defined above should be used only if the actual compatibility relation is not available. It is understood that the compatibility relation defined by equation (3.1) does not necessarily represent faithfully the actual compatibility relationship.

In general the underlying probability functions on $P(\{s\})$ and $P(\{s'\})$ alone are not sufficient to completely describe the joint probability function $P(\{(s, s')\})$, which however must satisfy the following constraints:

$$P(\{s'\}) = \sum_{s \in S} P(\{(s, s')\}), \text{ and } P(\{s\}) = \sum_{s' \in S'} P(\{(s, s')\}) \text{ for all } s' \in S' \text{ and } s \in S. \quad (3.2)$$

In addition to these constraints, $P(\{(s, s')\})$ must also satisfy:

$$P(\{(s, s')\}) = 0 \text{ if } (s, s') \text{ is not compatible with any } t \in T, \quad (3.3)$$

as imposed by the compatibility relation $C \oplus C'$. The constraints defined by equation (3.3) are necessary to ensure that the function $m_{S \times S'}$:

$$m_{S \times S'}(F) = \sum_{\{(s, s')\} \Rightarrow F} P(\{(s, s')\}).$$

is a basic probability assignment as defined by equation (2.1). Any probability function $P(\{(s, s')\})$ which satisfies constraints (3.2) and (3.3) is called an extension (Hartmanis, 1959) of the functions $P: 2^S \rightarrow [0, 1]$ and $P: 2^{S'} \rightarrow [0, 1]$. There may exist an infinite number of such extensions for a given pair of probability functions defined on $S$ and $S'$. If no such extension exists, we say that there is a fundamental disagreement or conflict between these two functions. Since there are many possible extensions for a given non-conflicting pair of probability functions, one will have to decide on the most appropriate joint probability function. In the Bayesian approach, the conditional probabilities enable us to determine the actual $P(\{(s, s')\})$ using the Bayes rule of conditionalization:

$$P(\{(s, s')\}) = P(\{s\}) \cdot P(\{s'\} \mid \{s\}) \text{ for all } s' \in S' \text{ and } s \in S, \quad (3.4)$$

where $P(\{s'\} \mid \{s\})$ is the probability that $s'$ is the true answer to the question which defines $S'$ given that $s$ is the true answer to the question which defines $S$.

The accuracy of the Bayesian approach will obviously depend on how accurately one can estimate the conditional probabilities $P(\{s'\} \mid \{s\})$ for all $s' \in S'$ and $s \in S$. These conditional probabilities must at the same time satisfy the constraints (3.2) and (3.3). Thus, in practice it may not always be feasible to provide a reasonable estimation of the conditional probabilities as required by the Bayesian approach.

The Dempster rule on the other hand adopts a simple multiplication axiom to compute the joint probability function as:

$$P(\{(s, s')\}) = K \cdot P(\{s\}) \cdot P(\{s'\}), \quad (3.5)$$

where K is a normalization constant used to ensure that the resulting probability function obeys the constraints defined by equation (3.3). It should perhaps be emphasized that the Dempster rule does not guarantee that the resulting joint probability function obeys the constraints specified by equation (3.2). This means that the probability function $P(\{(s, s')\})$ is not necessarily consistent with the probability functions $P(\{s\})$ and $P(\{s'\})$. Consequently, the belief in some of the propositions after combining the evidence may actually be lower than the belief originally assigned by the individual bodies of evidence. The theory of belief functions is supposed to enhance belief in a given proposition as more evidence in favor of the proposition becomes available. In other words, the belief in a proposition should increase monotonically with the accumulation of evidence, but the Dempster rule does not guarantee such a monotonic increase. We believe that this is one of the main drawbacks of the Dempster rule of combination.

**Example 3.1:** Assume that for the problem considered in Example 2.1 we have some additional evidence which defines another bpa $m_{S'}$:

$$m_{S'}(\{t_2, t_3\}) = 0.7, \quad m_{S'}(\{t_3\}) = 0.2 \text{ and } m_{S'}(\{t_1, t_2, t_3\}) = 0.1. \quad (3.6)$$

The basic probability numbers for all other subsets of $T$ are zero. The corresponding belief function is denoted by $Bel_{S'}$. Using the same technique as shown in Example 2.1 we can construct the underlying evidence frame $S' = \{s'_1, s'_2, s'_3\}$. The compatibility relation $C'$ between the elements of $S'$ and $T$ is defined by:

$$s'_1 C' t_2, \quad s'_1 C' t_3, \quad s'_2 C' t_3, \quad s'_3 C' t_1, \quad s'_3 C' t_2, \quad s'_3 C' t_3. \tag{3.7}$$

According to equation (2.5), the probability function on $S'$ obtained from $m_{S'}$ is given by:

$$P(\{s'_1\}) = 0.7, \quad P(\{s'_2\}) = 0.2 \text{ and } P(\{s'_3\}) = 0.1. \tag{3.8}$$

Now we can compute the degrees of belief based on the accumulated evidence by combining the belief functions $Bel_S$ and $Bel_{S'}$ defined by the bpa's in equations (2.6) and (3.6). As mentioned at the beginning of this section, the first step in the combination of two belief functions is the construction of the compatibility relation $C \oplus C'$ between $S \times S'$ and $T$. If the actual compatibility relation $C \oplus C'$ is not known, we can construct the relation $C \oplus C'$ from equation (3.1):

$$(s, s') \, C \oplus C' \, t \text{ if } s \, C \, t \text{ and } s' \, C \, t,$$

using the compatibility relations $C$ and $C'$ specified by equations (2.7) and (3.7), respectively. The resulting relationships are:

$$\begin{array}{lll}
(s_1, s'_3) \, C \oplus C' \, t_1, & (s_2, s'_3) \, C \oplus C' \, t_1, & (s_1, s'_1) \, C \oplus C' \, t_2, \\
(s_1, s'_3) \, C \oplus C' \, t_2, & (s_2, s'_1) \, C \oplus C' \, t_2, & (s_2, s'_3) \, C \oplus C' \, t_2, \\
(s_2, s'_1) \, C \oplus C' \, t_3, & (s_2, s'_2) \, C \oplus C' \, t_3, & (s_2, s'_3) \, C \oplus C' \, t_3
\end{array} \tag{3.9}$$

According to equations (2.8), (3.2) and (3.8), the constraints on the joint probability function $P(\{(s, s')\})$ can be explicitly written as:

$$P(\{(s_1, s'_1)\}) + P(\{(s_1, s'_2)\}) + P(\{(s_1, s'_3)\}) = P(\{s_1\}) = 0.8 \tag{3.10}$$

$$P(\{(s_2, s'_1)\}) + P(\{(s_2, s'_2)\}) + P(\{(s_2, s'_3)\}) = P(\{s_2\}) = 0.2 \tag{3.11}$$

$$P(\{(s_1, s'_1)\}) + P(\{(s_2, s'_1)\}) + P(\{(s_3, s'_1)\}) = P(\{s'_1\}) = 0.7 \tag{3.12}$$

$$P(\{(s_1, s'_2)\}) + P(\{(s_2, s'_2)\}) + P(\{(s_3, s'_2)\}) = P(\{s'_2\}) = 0.2 \tag{3.13}$$

$$P(\{(s_1, s'_3)\}) + P(\{(s_2, s'_3)\}) + P(\{(s_3, s'_3)\}) = P(\{s'_3\}) = 0.1 \tag{3.14}$$

Since the pair $(s_1, s'_2)$ is not compatible with any $t \in T$, from equation (3.3) we have the additional constraint on the joint probability function:

$$P(\{(s_1, s'_2)\}) = 0. \tag{3.15}$$

The second step of the combination is to construct the joint probability function $P(\{(s, s')\})$. If we were to apply the Bayes rule, we need conditional probabilities. In the absence of the required conditional probabilities we may use the Dempster rule of combination instead.

The joint probability function $P^\oplus$ obtained by the Dempster rule is:

$$P^\oplus(\{(s_1, s'_1)\}) = 0.667, \quad P^\oplus(\{(s_1, s'_3)\}) = 0.048, \quad P^\oplus(\{(s_2, s'_1)\}) = 0.095,$$
$$P^\oplus(\{(s_2, s'_2)\}) = 0.166, \quad P^\oplus(\{(s_1, s'_2)\}) = 0.000, \quad P^\oplus(\{(s_2, s'_3)\}) = 0.024.$$

Using equation (2.4) this function $P^\oplus$ together with the compatibility relationships in equation (3.9) define the basic probability assignment $m_S \oplus m_{S'}$ for the combined evidence:

$$m_S \oplus m_{S'}(\{t_2\}) = 0.667, \quad m_S \oplus m_{S'}(\{t_3\}) = 0.048, \quad m_S \oplus m_{S'}(\{t_1, t_2\}) = 0.095,$$
$$m_S \oplus m_{S'}(\{t_2, t_3\}) = 0.166, \quad m_S \oplus m_{S'}(\{t_1, t_2, t_3\}) = 0.024,$$

which in turn define the combined belief function $Bel_S \oplus Bel_{S'}$. The symbol $\oplus$ indicates that the combination is achieved using the Dempster rule.

It can be easily verified that the above joint probability function $P^\oplus$ does not satisfy the constraints (3.10) - (3.14). As a result, the belief in some propositions actually goes down with the accumulation of evidence using the Dempster rule. For example,

$$Bel_S(\{t_1, t_2\}) = 0.8 \quad \text{while} \quad Bel_S \oplus Bel_{S'}(\{t_1, t_2\}) = 0.762. \tag{3.16}$$

That is, $Bel_S \oplus Bel_{S'}(\{t_1, t_2\}) < Bel_S(\{t_1, t_2\})$. Such a decrease in belief seems unreasonable, particularly because both bodies of evidence assign high plausibility (1 and 0.9, respectively) to the proposition $\{t_1, t_2\}$. $\square$

So far we have analyzed two different rules of combination, namely, the Bayes rule and the Dempster rule. Each rule has its own drawbacks. The Bayes rule requires conditional probabilities which may not be available in practice. The Dempster rule on the other hand generates a joint probability function which may not necessarily be consistent with the individual probability functions. Is it possible to remove these drawbacks and construct a unified rule of combination? The following section introduces the principle of minimum information gain. We believe that this principle provides a possibility of unifying the Bayes and Dempster rules within the framework of information theory.

## 4. The Principle of Minimum Information Gain

According to information theory, the information contained in a probability function $P : 2^S \to [0, 1]$ defined on a frame of discernment $S$ can be expressed as (Lewis, 1959):

$$I_P(S) = \log |S| - H_P(S) ,\qquad(4.1)$$

where

$$H_P(S) = - \sum_{s \in S} P(\{s\}) \cdot \log P(\{s\}) \qquad(4.2)$$

is the entropy function of $P$, $|S|$ is the cardinality of frame $S$, and $\log |S|$ is the maximum entropy. (Since we are dealing with only *one* probability function here, the subscript $P$ in equation (4.1) can be ignored.)

In evidential reasoning, there is no *a priori* information about the truth value of any of the propositions. It can be easily seen that a probability function (distribution) conveys more information about the truth values of the propositions when it is more peaked and less information when it is uniform. The information measure $I(S)$, defined as the difference between the maximum entropy $\log |S|$ and the actual entropy $H(S)$, indeed reflects this property. If $P(\{s\}) = \dfrac{1}{|S|}$ for all $s \in S$, then $I(S) = 0$. If $P(\{s\}) = 1$ for some $s \in S$, then $I(S) = \log |S|$.

Let $S$ and $S'$ be two evidence frames for which the underlying probability functions are known based on two distinct bodies of evidence. According to equation (4.1), the information contained in $P : 2^S \to [0, 1]$ is represented by $I(S)$ and the information contained in $P : 2^{S'} \to [0, 1]$ is $I(S')$. Suppose the combination of these two bodies of evidence results in a joint probability function $P : 2^{S \times S'} \to [0, 1]$. Then the information provided by the combined evidence is $I(S \times S')$. The information gain $\Delta I(S, S')$ due to the combination can therefore be expressed as:

$$\Delta I(S, S') = I(S \times S') - I(S) - I(S') .\qquad(4.3)$$

The available information contained in the probability function $P(\{(s, s')\})$ on $S \times S'$ can be explicitly stated in terms of constraints. Equations (3.2) and (3.3) are examples of such constraints. If one has information about the probability function $P(\{(s, s')\})$ such as the conditional probabilities $P(\{s'\} | \{s\})$ for some $s \in S$ and $s' \in S'$, the additional constraints can be expressed as:

$$P(\{(s, s')\}) = P(\{s\}) \cdot P(\{s'\} | \{s\}) ,\qquad(4.4)$$

where $P(\{s\})$ and $P(\{s'\} | \{s\})$ are known quantities. Note that if $S$ and $S'$ are abstract frames we may not be able to define any conditional probabilities at all.

We have expressed all the available information in the form of constraints. Constraints defined by equation (3.2) represent the information we have about the probability functions defined on the frames $S$ and $S'$, respectively. Constraints given by equation (3.3) represent our knowledge about the joint compatibility relation $C \oplus C'$, while constraints (4.4) represent the information provided by the available conditional probabilities. If the joint probability function is consistent with the constraints (3.2), (3.3) and (4.4), it represents all the available information. As mentioned before, there exists many joint probability functions which represent all the available information. Which one of these functions should be adopted as the appropriate joint probability function? In this study we suggest that it is more appropriate to choose the joint probability function with the minimum information gain. Choosing any other joint probability function will result in more information gain than what can be justified by the available information. This means that the information gain from the combination process should be *minimal*.

According to equation (4.1), the information gain $\Delta I(S, S')$ can be written as:

$$\Delta I(S, S') = I(S \times S') - I(S) - I(S')$$
$$= H(S) + H(S') - H(S \times S') + \log |S \times S'| - \log |S| - \log |S'|$$
$$= H(S) + H(S') - H(S \times S') + \log \left[ \frac{|S \times S'|}{|S| \cdot |S'|} \right].$$

Since $|S \times S'| = |S| \cdot |S'|$, $\log \left[ \frac{|S \times S'|}{|S| \cdot |S'|} \right] = 0.$

$$\Delta I(S, S') = H(S) + H(S') - H(S \times S'). \tag{4.5}$$

From equations (4.2) and (4.5), we obtain:

$$\Delta I(S, S') = \sum_{(s, s') \in S \times S'} P(\{(s, s')\}) \cdot \log P(\{(s, s')\}) + \sum_{s \in S} P(\{s\}) \cdot \log P(\{s\})$$
$$+ \sum_{s' \in S'} P(\{s'\}) \cdot \log P(\{s'\})$$
$$= \sum_{(s, s') \in S \times S'} P(\{(s, s')\}) \cdot \log P(\{(s, s')\}) + \sum_{(s, s') \in S \times S'} P(\{(s, s')\}) \cdot \log P(\{s\})$$
$$+ \sum_{(s, s') \in S \times S'} P(\{(s, s')\}) \cdot \log P(\{s'\})$$
$$= \sum_{(s, s') \in S \times S'} P(\{(s, s')\}) \cdot \log \left[ \frac{P(\{(s, s')\})}{P(\{s\}) \cdot P(\{s'\})} \right] \tag{4.6}$$

Therefore, minimizing the information gain $\Delta I(S, S')$ is equivalent to minimizing the quantity:

$$H(P, P \times P) = \sum_{(s, s') \in S \times S'} P(\{(s, s')\}) \cdot \log \left[ \frac{P(\{(s, s')\})}{P \times P(\{(s, s')\})} \right].$$
$$= \sum_{(s, s') \in S \times S'} P(\{(s, s')\}) \cdot \log \left[ \frac{P(\{(s, s')\})}{P(\{s\}) \cdot P(\{s'\})} \right]. \tag{4.7}$$

The probability function $P \times P$ is defined by:

$$P \times P(\{(s, s')\}) = P(\{s\}) \cdot P(\{s'\}) \tag{4.8}$$

in which the probability functions on $S$ and $S'$ are assumed to be stochastically independent. The quantity $H(P, P \times P)$ is referred to as the cross-entropy of the probability function $P(\{(s, s')\})$ relative to $P \times P(\{(s, s')\})$ (Shore and Johnson, 1980). Sometimes, $H(P, P \times P)$ is called the mutual information (Hamming, 1980).

In general, the cross-entropy may be viewed as a measure of the closeness between a probability function $P_2$ defined on a frame $U$ and another probability function $P_1$ defined on the same frame. Under such an interpretation, we may express the cross entropy as:

$$H(P_1, P_2) = \sum_{u \in U} P_2(u) \cdot \log \left[ \frac{P_2(u)}{P_1(u)} \right]. \tag{4.9}$$

which is also known as the Kullback divergence (Kullback and Leibler, 1951). Suppose a prior probability function $P_1$ describes our belief based on the initial knowledge. Assume that some additional knowledge becomes available, which specifies a new set of constraints. In this case we will have to select a posterior probability function $P_2$ which obeys these constraints. Furthermore, the posterior function $P_2$ should be as close as possible to the prior function $P_1$. This means that the cross entropy $H(P_1, P_2)$ should be minimized under the given constraints. The minimum cross-entropy formalism, an extension of probability theory, has been used for estimating probabilities when very little information is available to allow application of classical methods (Shore and Johnson, 1980). It is a method of translating fragmentary probability information into a complete probability assignment.

From equations (4.7) and (4.9) it can be seen that the quantity $H(P, P \times P)$ is a special kind of cross-entropy with $U = S \times S'$, $P_1 = P \times P$, and $P_2 = P : 2^{S \times S'} \to [0, 1]$. That is, the principle of minimum information gain is a special case of the minimum cross-entropy formalism. As mentioned before, the function $P \times P$ defined by equation (4.8) is based on the assumption that the probability functions on $S$ and $S'$ are stochastically independent. In other words, the principle of minimum information gain favors the probability function which is closest to the one obtained by assuming stochastic independence.

In addition to the compatibility relations $C$, $C'$, $C \oplus C'$, we also need to know the underlying probability functions on $S$ and $S'$ for the combination. The quantities $H(S)$ and $H(S')$ defined by these probability functions are in fact constants. Thus, we can express the information gain $\Delta I(S, S')$ given by equation (4.5) as:

$$\Delta I(S, S') = -H(S \times S') + constant .$$

This means that minimizing the information gain $\Delta I(S, S')$ is equivalent to maximizing the entropy $H(S \times S')$.

Similar to the principle of minimum cross-entropy, the maximum entropy formalism is an extension of probability theory which has received considerable attention (Jaynes, 1957; Cooper and Huizinga, 1982; Tribus, 1969). In fact it can be shown that the principle of minimum cross-entropy is a generalization of the maximum entropy formalism (Shore and Johnson, 1980). The maximum entropy specifies the *maximally non-committal*, or *minimally prejudiced* probability function consistent with the given constraints (Tribus, 1969).

The above comments provide strong support for the principle of minimum information gain. This principle is equivalent to the minimum cross-entropy formalism when the prior probability function assumes stochastic independence. Thus, the joint probability function obtained by minimizing the information gain under the given constraints is as close to the stochastic independence case as possible. Since minimizing the information gain is also equivalent to maximizing the entropy, the resulting joint probability function can be considered as being the *maximally non-committal*, or *minimally prejudiced* under the given constraints.

There are many numerical methods for calculating the joint probability functions with minimum cross-entropy or maximum entropy (Brown, 1959; Cooper and Huizinga, 1980). A number of these techniques have been implemented and such programs are available. The following example illustrates the minimization using Lagrange multipliers.

**Example 4.1:** It can be easily verified that many joint probability functions satisfy the constraints (3.10) - (3.15) in Example 3.1. We will choose the most appropriate joint probability function by minimizing the information gain $\Delta I(S, S')$ using Lagrange multipliers. Let $\log k_1$ - $\log k_6$ be the Lagrange multipliers for equations (3.10) - (3.15). By differentiating $\Delta I(S, S') + \sum_{i=1}^{6} g_i \cdot \log k_i$ with respect to $P(\{(s, s')\})$ and equating the results to zero, we obtain:

$$\frac{\partial \Delta I(S, S')}{\partial P(\{(s, s')\})} + \sum_{i=1}^{6} \frac{\partial g_i}{\partial P(\{(s, s')\})} \cdot \log k_i = 0 \quad \text{for all } (s, s') \in S \times S',$$

where $g_1$ is the left side of equation (3.10), $g_2$ the left side of equation (3.11), and so on. Solving the above system of equations we arrive at the following values for $P(\{(s, s')\})$:

$$P(\{(s_1, s'_1)\}) = k_1 \cdot k_3, \quad P(\{(s_1, s'_2)\}) = k_1 \cdot k_4 \cdot k_6, \quad P(\{(s_1, s'_3)\}) = k_1 \cdot k_5,$$
$$P(\{(s_2, s'_1)\}) = k_2 \cdot k_3, \quad P(\{(s_2, s'_2)\}) = k_2 \cdot k_4, \quad P(\{(s_2, s'_3)\}) = k_2 \cdot k_5.$$

By substituting these values of $P(\{(s, s')\})$ into equations (3.10) - (3.15) and solving for $k_i$ we obtain the following joint probabilities:

$$P(\{(s_1, s'_1)\}) = 0.7, \ P(\{(s_1, s'_2)\}) = 0.0, \ P(\{(s_1, s'_3)\}) = 0.1,$$
$$P(\{(s_2, s'_1)\}) = 0.0, \ P(\{(s_2, s'_2)\}) = 0.2, \ P(\{(s_2, s'_3)\}) = 0.0 .$$

This joint probability function together with the compatibility relationships given by equation (3.9) between $S \times S'$ and $T$ immediately lead to the combined basic probability assignment:

$$m_{S \times S'}(\{t_2\}) = 0.7, \quad m_{S \times S'}(\{t_3\}) = 0.2, \quad m_{S \times S'}(\{t_1, t_2\}) = 0.1,$$

where the subscript $S \times S'$ emphasizes that the combination uses the probability function on $S \times S'$ obtained by minimizing the information gain. The combined basic probability numbers for all other subsets of $T$ are zero. Let the corresponding belief function be denoted by $Bel_{S \times S'}$. In contrast to the results obtained from the Dempster rule

(equation (3.16)), from our approach we obtain a combined belief function $Bel_{S \times S'}$ satisfying:

$$Bel_{S \times S'}(A) \geq Bel_S(A), \text{ and } Bel_{S \times S'}(A) \geq Bel_{S'}(A), \text{ for all } A \in 2^T.$$

These results indicate that our combination rule derived from the principle of minimum information gain indeed guarantees a monotonic increase in belief with accumulation of evidence. □

The method described in the last example provides a general solution for the minimization process. In what follows we discuss two special cases that have simple solutions, which clearly demonstrates our method to be a unification of the Bayes rule of conditionalization and the Dempster rule of combination.

*Case (i):* If all the conditional probabilities $P(\{s'\} | \{s\})$ are given, it is unnecessary to carry out the actual minimization. The following constraints as defined by equation (4.4):

$$P(\{(s, s')\}) = P(\{s\}) \cdot P(\{s'\} | \{s\}) ,$$

will completely describe the joint probability function $P(\{(s, s')\})$. Obviously, this special case is equivalent to the Bayes rule of conditionalization. There have been previous attempts of incorporating dependencies in the Dempster rule (Ruspini, 1986, Yen, 1989). It can be easily seen that the method proposed here is a generalization of these approaches.

*Case (ii):* If one does not have any a priori knowledge about $P(\{(s, s')\})$ (i.e., there are no constraints of the type given by equations (3.3) and (4.4)), the minimization is trivially satisfied by the following joint probability function:

$$P(\{(s, s')\}) = P(\{s\}) \cdot P(\{s'\}) .$$

This result is equivalent to the Dempster rule of combination when the normalization constant $K$ defined by equation (3.5) is equal to one.

It is interesting to note that our method based on the principle of minimum information gain is equivalent to the Bayes rule when all the necessary information is available. Our rule is equivalent to the Dempster rule of combination only when the normalization constant $K$ is equal to one. It should be noted that the proposed minimization procedure substantially differs from the Dempster rule when $K$ is not equal to one. The use of normalization constant in the Dempster rule is debatable. If $K \neq 1$, the two bodies of evidence being combined are said to be in conflict with one another (Shafer, 1976). Such a definition of conflict may not be reasonable. According to our approach if it is possible to construct a joint probability function on $S \times S'$ that is consistent with the individual probability functions on $S$ and $S'$ and the additional constraints given by equations (3.3) and (4.4), we do not have sufficient reason to assume that there is a conflict between the two bodies of evidence. We wish to emphasize here that the principle of minimum information gain does not presume a conflict between the two bodies of evidence if it is possible to construct a joint probability function consistent with the given constraints. (See Examples 2.1, 3.1 and 4.1). Such an approach is especially helpful in maintaining consensus among different bodies of evidence. Thus, with our framework if under certain circumstances it is not possible to satisfy all the given constraints, one can then say that there is a fundamental conflict between the available information. In this case, the available information may have to be reassessed based on the reliability of the individual evidence before procceding to minimization of the information gain. The resolution of conflicts can be achieved with the comparative belief structures (Wong and Lingras, 1990b) or by the discounting of belief functions (Shafer, 1973). It is perhaps worth mentioning that the principle of minimum information gain can be easily extended to combine more than two bodies of evidence (Wong and Lingras, 1990a).

## 5. Summary

With limited dependency information about the accumulated belief the Dempster rule of combination produces unsatisfactory results. The present study suggests a method to determine the accumulated belief based on the premise that the information gain from the combination process is minimum. The proposed principle of minimum information gain is a special case of the principle of minimum cross-entropy, when the prior probability is obtained by assuming stochastic independence. This means that the probability function obtained from minimizing the information gain is as close to the stochastic independence case as possible under given constraints. Since we have also shown that minimum information gain corresponds to maximum entropy, the resulting probability function can be considered maximally non-committal, or minimally prejudiced.

The principle of minimum information gain enables us to incorporate available conditional probabilities in the combined belief function. This is a generalization of previous attempts to incorporate dependencies in the Dempster

459

rule. We have also demonstrated that the Bayes and Dempster rules can be viewed as special cases of our rule of combination. Finally, we show that the application of the principle of minimum information gain does result in monotonic increase in belief with accumulation of consistent evidence. Our method may provide a more reasonable criterion for identifying conflicts among different bodies of evidence for approximate reasoning.